\documentclass[11pt,a4paper]{article}
\usepackage[hyperref]{naaclhlt2019}
\usepackage{times}
\usepackage{latexsym}
\usepackage{graphicx}  %Required
\usepackage{amsfonts}
\usepackage{bm}
\newcommand{\argmax}{\mathop{\rm arg~max}\limits}

\usepackage{eqparbox}
\usepackage{arydshln}
\usepackage{CJKutf8}
\usepackage[multiple]{footmisc}
\usepackage{multirow}
\usepackage{todonotes}
\usepackage{amssymb}  
\usepackage{amsthm}
\usepackage{amsmath}

\newcommand*{\Ja}[1]{%
   \begin{CJK}{UTF8}{ipxm}{\small #1}\end{CJK}} % {\small }使うと日本語表記のサイズが1p下がる

\usepackage{url}

\title{Multilingual Extractive Reading Comprehension \\
by Runtime Machine Translation}

\author{Akari Asai$^\dagger$,  Akiko Eriguchi$^\dagger$, Kazuma Hashimoto$^\ddagger$, and Yoshimasa Tsuruoka$^\dagger$\\
$^\dagger$The University of Tokyo~~~~~~$^\ddagger$Salesforce Research\\
$^\dagger${\tt akari-asai@g.ecc.u-tokyo.ac.jp}\\
$^\ddagger${\tt \{eriguchi,tsuruoka\}@logos.t.u-tokyo.ac.jp}\\
$^\ddagger${\tt k.hashimoto@salesforce.com}
}

\date{}

\begin{document}
\maketitle
\begin{abstract}
Despite recent work in Reading Comprehension (RC), progress has been mostly limited to English due to the lack of large-scale datasets in other languages.
In this work, we introduce the first RC system for languages without RC training data.
Given a target language without RC training data and a pivot language with RC training data (e.g. English),
our method leverages existing RC resources in the pivot language by combining a competitive RC model in the pivot language with an attentive Neural Machine Translation (NMT) model.
We first translate the data from the target to the pivot language, and then obtain an answer using the RC model in the pivot language.
Finally, we recover the corresponding answer in the original language using soft-alignment attention scores from the NMT model.
We create evaluation sets of RC data in two non-English languages, namely Japanese and French, to evaluate our method.
Experimental results on these datasets show that our method significantly outperforms a back-translation baseline of a state-of-the-art product-level machine translation system. 
% Moreover, our analysis suggests that the ability to translate question sentences is particular important for RC in non-English languages task, and that oversampling a small number of manually translated questions in addition to an automatically created corpus is critical to attaining good performance.
\end{abstract}

\section{Introduction}
\begin{figure*}[!h]
  \centering
  \includegraphics[width=\textwidth]{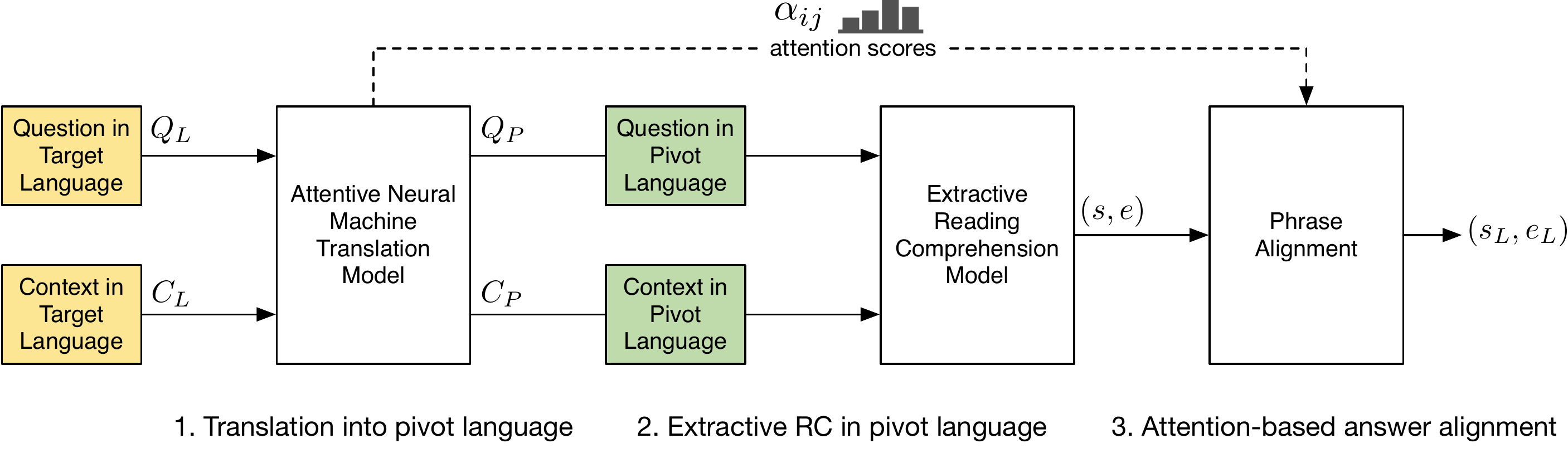}
  \caption{Overview of our method. $\bm{\alpha}_{ij}$ are the attention weights~(attention distribution) in the NMT model. $(s, e)$ and $(s_\mathrm{L}, e_\mathrm{L})$ are the answer spans in the pivot language (e.g. English) and target language $L$, respectively.}
  \label{fig:overview}
\end{figure*}
Extractive Reading Comprehension (RC), in which a model identifies the answer to a given question from a document context by ``extracting'' the correct answer, has a variety of downstream applications such as search, automated FAQs, and dialogue systems.
%Extractive Reading Comprehension (RC) is a Question Answering (QA) task to identify the answer to a given question from the text of a given paragraph, where a system is required to ``extract'' the correct answer to a question from the reference paragraph.
Recent years have seen rapid progress in the development of RC models~\cite{seo2016bidirectional,wang-EtAl:2017:Long2,xiong2017dcn+,yu2018qanet,hu2017mnemonic} due to the availability of large-scale annotated corpora~\citep{nips15_hermann,D16-1264,P17-1147}.
However, these large-scale annotated datasets are often exclusive to English.
Consequently, progress in RC has been largely limited to English.

To alleviate the scarcity of training data in non-English languages, previous work creates a new large-scale dataset for a language of interest~\cite{he2017dureader} or combines a medium-scale dataset in the language with an existing dataset translated from English~\cite{lee2018semi}.
These efforts in data creation are often costly, and must be repeated for each new language of interest.
In addition, they do not leverage existing resources in English RC, such as the wealth of large-scale datasets and state-of-the-art models.

In this paper, we propose a multilingual extractive RC method by runtime Machine Translation (MT), a new method for building RC systems for languages without RC training data.
Our method combines an RC model with a Neural Machine Translation model (NMT).
Given a language $L$ of interest with no RC data and a pivot language $P$ with large-scale RC training data, we first translate a document context and question from the language $L$ to the language $P$ using an attentive NMT model.
Next, we obtain an answer from the RC model in language $P$.
Finally, we recover the answer in the context in language $L$ using soft-alignments from the NMT model.

% how do we address the shortcoming of existing methods
To our knowledge, our work is the first method that requires no RC training data in the target language to build an RC model for the language of interest.
In contrast to existing work on RC in non-English languages, our method leverages existing work in English RC.
More importantly, our method requires no additional annotation effort to acquire RC data in the target language.
% That is, our method is applicable to any language as long as the NMT model for the target language can be performed with a certain level of accuracy, 
% That is, given the availability of the machine translation model, our approach does not require extra data annotation effort and scales to additional languages.

To demonstrate the effectiveness of our method, we focus on SQuAD, one of the most widely-used large-scale English datasets for extractive RC, and created SQuAD test data in Japanese and French.
% We demonstrate the effectiveness of our method on a French evaluation set and a Japanese evaluation set, focusing on Stanford Question Answering Dataset (SQuAD)~\citep{D16-1264}.  
% We curate both of these evaluation sets, which we refer to as multilingual SQuAD, following the extractive question answering style of SQuAD.
On Japanese and French SQuAD, our method significantly outperforms a back-translation baseline that first translates from the target language $L$ to the pivot language $P$, produces an answer in language $P$, and back-translates the answer into the language $L$.
% Our method achieves this result despite using an inferior MT model (an attentive sequence-to-sequence model) compared to that of the baseline (Google Translate).
%Our method achieves this result despite using an MT model trained on fewer resources than that one used in the back-translation baseline.
Our method achieves this result despite using much smaller translation data than that of a state-of-the-art MT system used in the back-translation baseline.

Analysis of our experiments shows that the ability to correctly translate questions is crucial for the end task of RC.
In particular, oversampling a small set of high quality question translations in training the NMT model results in significant accuracy gains in RC.
Moreover, our error analysis shows that under-translation and paraphrasing in translation significantly degrade the downstream RC accuracy, although they do not have large effects on BLEU scores.
% These results indicate that using the relevant NMT components in a task-specific manner is more important than improving the BLEU scores. 
% \todo{remember to anonymize this link when submitting to NAACL}
We make our code and the collected Japanese and French SQuAD datasets available at~\url{https://github.com/AkariAsai/extractive_rc_by_runtime_mt}.

\section{Extractive RC by Runtime MT}
\label{sec:method}

Given a language $L$ of interest with no RC training data and a pivot language $P$ with a copious amount of RC training data, our method leverages an attentive NMT model that translates from language $L$ into language $P$ and a RC model in language $P$.
For a document context and question in the language $L$, we first translate the context and question to the pivot language $P$ using the attentive NMT model.
Next, we obtain an answer in language $P$ using the RC model.
Finally, we recover the answer in $L$ using soft-alignments from the attentive NMT model.
Figure~\ref{fig:overview} provides an illustration of our method.
Here, we assume that we have a bilingual corpus for $P$ and $L$ with which to train the NMT model and an extractive RC dataset in the language $P$ with which to train the RC model.

\subsection{Translation to Pivot Language}
% keep this part here, as for the past multi-lingual QA works, many people first use Google Translate or Moses first, instead of their own MT models. 
To translate the the context and question from the target language $L$ into the pivot language $P$, one possible approach is to use a web service or a software package for MT (e.g. Google Translate\footnote{\url{https://translate.google.com/}}) 
as a blackbox MT system~\citep{Hartrumpf,esplagomis-sanchezmartinez-forcada:2012:STARSEM-SEMEVAL,W06-1906}.
However, this approach does not allow us to access the internal intermediate information that is potentially useful for bridging the MT and RC systems. 
% It should also be noted that the reproducibility of experiments is not guaranteed when using such off-the-shelf MT systems.

To overcome the limitations, we instead train an attention-based NMT model~\citep{luong-pham-manning:2015:EMNLP} as a white-box MT system.
Our attention-based NMT implementation uses an bidirectional recurrent neural network (RNN) encoder and a unidirectional RNN decoder with bilinear attention.
Given an input sentence of length $T$, we denote the hidden state of the encoder corresponding to the $i$-th word as $h_i\in\mathbb{R}^{d_1}$, where $d_1$ is the size of the encoder hidden state.
Similarly, we denote the hidden state of the decoder while generating the $j$-th output word as $\tilde{h}_j\in\mathbb{R}^{d_2}$, where $d_2$ is the size of the decoder hidden state.
We use bilinear attention, which computes the attention score $\alpha_{ij}$ between the $j$-th output word and the $i$th input word.

\begin{equation}
\alpha_{ij} = \frac{\exp{(h_{i}W{\tilde{h}_{j}})}}{\sum_{k=1}^{T}  \exp{(h_{k}W{\tilde{h}_{j}})}}.
\label{eq:attention}
\end{equation}

Here, $W$ is a parameter matrix.
The attention score $\alpha_{ij}$ estimates how informative $h_i$ is when predicting the $j$-th target word.

Let ${\rm MT}(X) \rightarrow Y$ denotes our MT model that translates a sequence $X$ to a sequence $Y$.
We translate the context $C_L$ and question $Q_L$ in $L$ to the corresponding context $C_P$ and question $Q_P$ in $P$.

\begin{eqnarray}
{\rm MT}(C_L) \rightarrow C_P,\\
{\rm MT}(Q_L) \rightarrow Q_P.
\end{eqnarray}

% and ${\rm QA}(Q, C) \rightarrow A$ denote the RC model that produces an answer $A$
% \subsection{Translation into English}
% To translate the input paragraphs and questions into English, one possible approach is using a web service or a software package for Machine Translation (MT) (e.g., Google Translate\footnote{\url{https://translate.google.com/}}) 
% as a black-box MT system~\citep{Hartrumpf,esplagomis-sanchezmartinez-forcada:2012:STARSEM-SEMEVAL,W06-1906}.
% \if0{
% as a black-box MT system as introduced in some previous work in multilingual question Answering (MLQA)~\citep{Hartrumpf,esplagomis-sanchezmartinez-forcada:2012:STARSEM-SEMEVAL,W06-1906,Trec98manualqueries}.
% }\fi
% However, this approach does not allow us to access the internally processed information that is potentially useful for bridging the MT and RC systems. 
% It should also be noted that the reproducibility of experiments is not guaranteed when using such off-the-shelf MT systems.
% \if0{
% We therefore implement an attention-based NMT model~\citep{luong-pham-manning:2015:EMNLP} as a whitebox MT system.
% }\fi
% 
% To overcome the limitations, we train an attention-based NMT model~\citep{luong-pham-manning:2015:EMNLP} as a whitebox MT system.

\subsection{Extractive RC in Pivot Language}

Having translated the question and context from the original language $L$ to the pivot language $P$, we now apply an RC model trained in the language $P$.
In this work, we use a variant of the Bidirecional Attention Flow model~\citep{clark2017simple,seo2016bidirectional}.
Similar models have proven successful on a variety of extractive RC tasks~\cite{seo2016bidirectional,wang-EtAl:2017:Long2,xiong2017dcn+,yu2018qanet,hu2017mnemonic}.

To identify an answer, the RC model outputs distributions corresponding to the start and end locations of the answer span in the context.
We denote these by $p_s(i)$ and $p_e(i)$, the probabilities of the $i$-th word being the start and end of the answer span.
We choose the span whose start position $s$ and end position $e$ yield the largest joint probability, given that the end position occurs after the start position (e.g. $s \le e$).

\begin{equation}
(s, e) = \argmax_{(m, n),~m\leq n} p_s(m) p_e(n).
\label{eq: se}
\end{equation}

\subsection{Answer Alignment to the Target Language}
\label{subsec:alignment}
Having produced the answer in the pivot language $P$ using the RC model, how do we find the corresponding answer in the language $L$?
One approach is to back-translate the answer using another $P$-to-$L$ MT system.
However, we find that directly translating the answer from the pivot language $P$ tends to yield irrelevant answers in the language $L$ because the model lacks grounding from the context and question in $L$.

We instead propose a method to align the start and end positions of the answer in the language $P$ with a span of the context $C_L$ in the language $L$ using attention weights $\alpha$ from equation~\eqref{eq:attention}.
As shown by~\citet{bahdanau+al-2014-nmt,luong-pham-manning:2015:EMNLP}, NMT attention weights $\alpha_{ij}$ provide an estimate of how informative $h_i$ is in predicting the $j$-th target word.
Consequently, we recover the answer in the original language by aligning each $j$-th word in the pivot language answer with its corresponding word $\ell(j)$ in the language $L$ context.
For a position $j$ in the pivot language context $C_P$, we choose $\ell(j)$, the corresponding position in the target language context $C_L$, as follows.

\begin{equation}
\ell(j) = \argmax_{i,~1\leq i \leq T} \alpha_{ij}.
\end{equation}

Given an answer in the pivot language demarcated by positions $(s, e)$ in the pivot language context $C_P$,
we recover the corresponding target language answer by choosing the largest aligned span.
Let $(s_L, e_L)$ denote the positions of the answer in the target language context $C_L$.
We compute $(s_L, e_L)$ as

\begin{eqnarray}
s_L &=& \min \{ \ell(s), \ell(s+1), \cdots \ell(e) \},\\
e_L &=& \max \{ \ell(s), \ell(s+1), \cdots \ell(e) \} .
\end{eqnarray}

\section{Japanese and French SQuAD}
\label{sec:test_dataset}
To evaluate the effectiveness of our method, 
% we newly create a Multilingual Question Answering task in Japanese and in French, by manually translating the original SQuAD dataset. 
we create SQuAD test datasets in Japanese and French to evaluate the proposed system. 
% need to good expression...?
Because the test set of SQuAD is not publically available, we instead create parallel examples from the SQuAD development dataset.
These parallel data are used solely for evaluation.
That is, no training data in the language is used by the RC model.

The SQuAD development set contains 2,067 paragraphs over 48 articles for a total of 10,570 paragraph-question pairs.
% To ensure diversity of questions in our test sets\todo{i dont understand why this encourages diversity}, 
We extract the first paragraph and its corresponding questions, resulting in 327 paragraph-question pairs over 48 articles.
The paragraphs and questions are then manually translated into each target language by bilingual workers on Amazon Mechanical Turk\footnote{\url{https://www.mturk.com/}}.
These translations are subsequently verified by bilingual experts, who also extract the corresponding answers in the translated paragraph, making sure that the answer in the target language retains the same meaning and context as the answer in English.

\section{Experiments in MT}
% We first present experimental results of our \{Japanese, French\}-to-English NMT model's translation performance on Wikipedia paragraph and question sentences.
We now describe our results in MT, which produces the attentive NMT model to translate the question and context from the target language $L$ to the pivot language $P$.
For experimental setup, please see Section C of the Appendix. 

\subsection{L-to-P Bilingual Corpus for SQuAD}
\label{subsec:wiki_enja}
One idea to train the NMT model is to use an existing parallel corpus.
However, in our preliminary experiments, we find that even a state-of-the-art Japanese-to-English NMT model trained with the ASPEC corpus~\citep{NAKAZAWA16.621}, an established English-Japanese corpus, results in poor performance on Japanese SQuAD.
This is because the training domain (scientific articles) differ considerably from the inference domain (Wikipedia).
The vocabulary and writing style in ASPEC is biased toward scientific fields and tend to be abstruse, while the Wikipedia-based SQuAD dataset covers domains such as musical celebrities and abstract concepts~\citep{D16-1264} with a generally simple writing style.

To address this domain mismatch, we construct new Wikipedia-based bilingual corpora using a sentence aligner\footnote{\url{https://github.com/danielvarga/hunalign}} on the Japanese and French Wikipedia articles and their English counterparts.
Finally, we select the 1,002,000 best aligned sentence pairs, and split the pairs into a training dataset of 1,000,000 pairs and a development dataset of 2,000 pairs. 
More details can be found in Section A of the Appendix.
This data is used to train our NMT model.

\subsection{Translations of Question Sentences} 
\label{subsec:question_translation}
Preliminary experiments on MT show that our NMT models tended to fail to translate question sentences due to data imbalance.
This is due to reasons noted by~\citet{D16-1163}, namely that NMT requires a copious amount of data to generalize, and learns poorly from low-count events.

We observe that question sentences are contained in 0.1\% of the Wikipedia-based bilingual corpus.
Moreover, most of the sentences are movie titles, captions, and quotes, which differ from SQuAD-style question sentences.
% \todo{I don't understand the part that follows}
% \citet{DBLP:journals/corr/HuPQ17} and \citet{zhang2017exploring} find that interrogatives play an important role in SQuAD models, 
% We observed that the question sentences were translated as declarative sentences, by feeding the low-quality question translated by the NMT model trained on .
Therefore, we introduce the following two approaches to address the problem of low-quality question translations.

\paragraph{Adding Manually Translated Questions}
We randomly sample 200 questions in English from the SQuAD training set, manually translate them into the language of interest $L$, and add these translated questions to the bilingual corpus.
More details can be found in the supplementary material Section B.

\paragraph{Oversampling}
\citet{chu-dabre-kurohashi:2017:Short,TACL1081} show that in a domain adaptation setting, NMT performance on a low-resource domain can be improved by oversampling a small corpus in the target domain.
We adapt this idea by oversampling the above-mentioned manually translated questions to emphasize these high quality translations during training.
We first duplicate the manually translated question sentences $l$ times, then mix the duplicated questions with the bilingual corpus.

\subsection{Results}
\label{results_MT}

% test by hashimoto
\begin{table}[t!]
\begin{center}
{\small
\begin{tabular}{|l|c|c|c|c|}
\hline
 & \multicolumn{2}{c|}{Ja-En} & \multicolumn{2}{c|}{Fr-En} \\
Translation method  & \scriptsize{\bf Wiki} & \scriptsize{\bf Question} & \scriptsize{\bf Wiki} & \scriptsize{\bf Question} \\ \hline
Our NMT & 23.95  & 22.75 & 45.64 & 40.47 \\
\hdashline
Google Translate &  24.09 &  37.98 & 41.08 & 50.91 \\
\hline
\end{tabular}
}
\end{center}
\caption{\label{Development BLEU}  MT BLEU scores of \{Japanese, French\}-to-English NMT on the bilingual Wikipedia development dataset (Wiki) and SQuAD question sentences (Question).} 
\end{table}

Our NMT model achieves 23.95 BLEU on Japanese-to-English and 45.64 on French-to-English, as shown in the ``Wiki'' column of Table~\ref{Development BLEU}.
Table~\ref{Development BLEU} also shows BLEU scores of translations produced by Google Translate\footnote{The translations were obtained at \url{https://translate.google.com} in August, 2018.}.
The competitive results attained by our NMT model compared to that of Google Translate, a state-of-the-art MT system, indicate that our proposed technique of automatically collecting parallel corpora is effective.

As shown in the columns of ``Question'' in Table~\ref{Development BLEU}, the BLEU scores of our NMT models for the question sentences are significantly lower than those of Google Translate.
This is not surprising; Google Translate is trained on their internal corpora which are three or four orders of magnitudes larger than our training corpora~\citep{TACL1081}.
Therefore, one promising research direction is to focus on how to further improve translation accuracy of question sentences, because question translations are crucial for our task.

\subsection{Ablation Study}

%revised by asai
\begin{table}[t!]
\begin{center}
{\small
\begin{tabular}{|l|c|c|c|c|}
\hline
 & \multicolumn{2}{c|}{Ja-En} & \multicolumn{2}{c|}{Fr-En} \\
Translation method  & \scriptsize{\bf Wiki} & \scriptsize{\bf Question} & \scriptsize{\bf Wiki} & \scriptsize{\bf Question} \\ \hline
% Add beam search?
Our NMT & 23.95  & 22.75 & 45.64 & 40.47 \\
w/o beam search & 20.78 & 23.06 & 41.93 & 36.21 \\
w/o question  &  \multirow{2}{*}{20.76}  & \multirow{2}{*}{16.94} & \multirow{2}{*}{42.05} & \multirow{2}{*}{35.03} \\ 
\ \ \ \ \ \ \ oversampling & &&& \\
w/o questions  &  20.36  & 10.68 & 41.37 & 22.75 \\ 
\hline
\end{tabular}
}
\end{center}
\caption{\label{table:Ablation_study} MT ablation study on \{Japanese, French\}-to-English translation showing the development set BLEU score.
The ablations are
1) removing beam search.
2) using manually translated questions without oversampling them.
3) not using manually translated questions.
}
\end{table}

% \todo{add ablation study for attention-alignment method}
In Table~\ref{table:Ablation_study}, we report an ablation study for our NMT models.
As a standard practice, beam search is effective in improving the BLEU scores, except for the Japanese-to-English question translations.
More importantly, the use of the small amount of manually translated questions is crucial.
Without using the question translations as shown in the row of ``w/o questions'', the BLEU scores are almost halved.
Our oversampling technique for the question translations is effective in better translating the question sentences.

Although our BLEU scores are competitive with the Google Translate results for the Wikipedia sentence translations, the BLEU scores of the Japanese-to-English dataset are much lower than those of the French-to-English dataset.
To reduce the gap, we experimented with jointly using the external parallel corpus, ASPEC, for training our Japanese-to-English model, and the ``Wiki'' BLEU score slightly improved from 23.95 to 24.47.
However, the ``Question'' BLEU score dropped from 22.75 to 21.80.
These results suggest that jointly using a high-quality external corpus does not always lead to a major improvement of translation performance.  
Moreover, it once again indicate the importance of questions translation pairs, which rarely occur in existing MT corpora.
Because we did not observe significant improvements, we do not rely on the external corpus in our main experiments.
% \todo{I'm not sure how the two clauses of this sentence, which form a conjunction, relate to each other}
% \todo{using the external corpora would limit the applicability to another language, and also the performance do not that improve even we use it, so we do not use external corpora. think about better and clear explanation}

\section{Experiments in RC}
\label{results_SQuaD}

% test by hashimoto 3
\begin{table}[t!]
\begin{center}
{\small
\begin{tabular}{|l|c|c|c|c|}
\hline
 & \multicolumn{2}{c|}{Japanese} & \multicolumn{2}{c|}{French} \\
Method  & \scriptsize{\bf F1} & \scriptsize{\bf EM} & \scriptsize{\bf F1} & \scriptsize{\bf EM} \\ \hline
Our method & {\bf 52.19} & {\bf 37.00} & {\bf  61.88 } & {\bf 40.67} \\\hdashline
Back-translation by & \multirow{2}{*}{42.60} &  \multirow{2}{*}{24.77} & \multirow{2}{*}{44.02} & \multirow{2}{*}{23.54} \\
using Google Translate & & & & \\ 
% Back-translation by & \multirow{2}{*}{42.60} &  \multirow{2}{*}{24.77} & \multirow{2}{*}{44.02} & \multirow{2}{*}{23.54} \\
% using Google Translate & & & & \\ 
% \hline
% Beam search & 23.95 \\ 
\hline
\end{tabular}
}
\end{center}
\caption{\label{squad_result} RC results of our method and the baseline on Japanese and French SQuAD.
The BiDAF model trained on the original English SQuAD dataset achieves an F1 score of 77.1 and an EM score of 67.2.} 
\end{table}

We now describe our results on the Japanese and French SQuAD tasks, using the best NMT model from Section~\ref{results_MT}.
For experimental details, please see Section E of the Appendix.
Our BiDAF model achieves an F1 score of 77.1 and an EM score of 67.2, while our BiDAF + Self Attention + ELMo achieves an F1 score of 83.2 and EM score of 74.7 on the SQuAD v1.1 English development dataset.
To compare our method to the baseline, we use the BiDAF + Self Attention + ELMo model~\cite{clark2017simple,peters2018deep} as our RC model.

\subsection{Baseline}
% add expalnation about the baseline, and why there are no strong comparison. 
As this is the first work to build an extractive RC system for a new language with no training data in the target language, there is no directly comparable approach. 
Previous work in Multilingual Question Answering (MLQA) is not easily applicable to multilingual RC due to differences in formulation as mentioned in Section~\ref{sec:related_work}.

Instead, we use a simple baseline, which employs a production translator (Google Translate) for both $L$-to-$P$ and $P$-to-$L$ translation. 
First, we translate the target language context $C_L$ and question $Q_L$ using the $L$-to-$P$ production translator. 
Next, the translated question $Q_P$ and context $C_P$ in the pivot language are given to the RC model to identify the answer in the pivot language $P$.
Finally, we use the $P$-to-$L$ production translator to back-translate the predicted answer from the pivot language $P$ to the target language $L$.
We refer to this baseline as ``{\it back-translation} system''. 

% \subsection{Experimental Settings}
% We train BiDAF~\citep{seo2016bidirectional} and BiDAF + Self Attention + ELMo model~\citep{clark2017simple} with the original SQuAD v1.1 English training dataset.
% For the training of the BiDAF model, we follow the training settings in \citet{seo2016bidirectional}, except that we set the batch size to 40 and trained the model for 20 epochs.
% In case of training the BiDAF + Self Attention + ELMo model, we follow the settings shown in \citet{clark2017simple}, adding ELMo embeddings~\cite{peters2018deep}.
% The only distinction is that we use 100 dimensional pre-trained GloVe~\cite{pennington2014glove}, instead of the 300 dimensional one. 
% To evaluate the performance, we calculate EM, which measures the exact match with the ground truth answers, and F1, the weighted average of the precision and recall rate at character level~\citep{D16-1264}.
% The initial learning rate is set to 0.5, and the learning rate was halved when no improvement was seen in the EM score during two epochs. 

\subsection{Result}
Table~\ref{squad_result} compares the performance of our method against the back-translation baseline on the Japanese and French SQuAD tasks.
Our method achieves the best F1 score of 52.19 and the best EM score of 37.00 on Japanese SQuAD, and the best F1 score of 61.88 and the best EM score of 40.67 on French SQuAD.
Our best model outperforms the back-translation baseline by 9.59 F1 and 12.23 EM on Japanese SQuAD, and by 17.86 F1 and 17.13 EM on French SQuAD.
The results on two very different languages suggest that our method is potentially applicable to a variety of languages.
In addition, we note that we outperform the baseline despite the latter using Google Translate, which performs considerably better than our NMT model in terms of BLEU scores as shown in Table~\ref{Development BLEU}.
These results underline the importance of using soft-alignments from a white-box attentive NMT model, as opposed to a more performant black-box translation system that obtains higher BLEU score.

\subsection{Ablation Study}

% test by hashimoto 3
\begin{table}[t!]
\begin{center}
{\small
\begin{tabular}{|l|c|c|c|c|}
\hline
 & \multicolumn{2}{c|}{Japanese} & \multicolumn{2}{c|}{French} \\
Method  & \scriptsize{\bf F1} & \scriptsize{\bf EM} & \scriptsize{\bf F1} & \scriptsize{\bf EM} \\ \hline
Our method &  52.19 &  37.00 & 61.88 &  40.67 \\
w/o self attention & \multirow{2}{*}{50.08}  & \multirow{2}{*}{35.47} & \multirow{2}{*}{57.56} & \multirow{2}{*}{37.61} \\ 
\ \ \ \ \ \ \ ELMo &&&& \\
w/o beam search  & 50.59 & 34.55 & 55.14 & 36.69 \\
w/o question & \multirow{2}{*}{33.97}  & \multirow{2}{*}{20.48} & \multirow{2}{*}{49.28} & \multirow{2}{*}{29.66} \\ 
\ \ \ \ \ \ \ oversampling &&&&\\
w/o questions  & 25.20  & 14.63 & 41.63 & 26.60 \\ 
\hline
\end{tabular}
}
\end{center}
\caption{\label{table:ab_squad} RC ablation results of our proposed method for the Japanese and French SQuAD task.
The four ablations are
1)
replacing the RC model~\citep{clark2017simple} with the base BiDAF model~\citep{seo2016bidirectional}.
2) removing beam search.
3) removing oversampling manually translated questions.
4) removing manually translated questions.
} 
\end{table}

Table~\ref{table:ab_squad} shows the ablation study of our method, which consists of our best Japanese-to-English and French-to-English NMT model and the BiDAF + Self Attention + ELMo model.

Similar to our finding in the MT ablation study, adding manually translated questions and oversampling are critical to the end task of RC.
On both of the Japanese and French SQuAD, we observed that NMT model trained only with the Wikipedia bilingual corpus (e.g. without manually translated questions) translated question sentences poorly --- the performance deteriorates by 26.99 F1 and 22.37 EM on Japanese, and by 20.25 F1 and 14.07 EM on French.
Question oversampling also significantly improves performance, especially on Japanese, where it increases F1 by 16.62 and EM by 14.47.
We find that our NMT models tend to translate the question sentences as declarative sentences when trained only on the Wikipedia bilingual corpora or when trained without oversampling of the manually translated questions.
For instance, a question ``\Ja{テスラは何年に亡くなったか。}(In what year did Tesla die?)'' is incorrectly translated as ``tesla died in a year .''.
This results in a distribution mismatch for the RC model, which has been trained with questions as opposed to declarative sentences.

Using a more competitive NMT or RC model is helpful, and combining both gives a notable improvement.
On the French SQuAD dataset, the self attention layer and ELMo in the extractive RC model improves the F1 by 4.32 and EM by 3.06, which are more than two times the improvements on the Japanese SQuAD dataset. 
We attribute the larger gain to the superior performance of the French-to-English NMT model on paragraph and question translation.

\subsection{Drawback of the Back-translation System}
We observe that the baseline back-translation method consistently makes mistakes due to its missing critical information from the context or question.
This lack of information often has a large negative impact on the extractive RC task, in which answers are expected to be precisely retrieved from the context. 
For instance, we observe instances in which the RC model specifies the correct answer span for the Japanese question ``\Ja{最初にアメリカ人を宇宙に送ったプログラムは何か？} (What project put the Americans into space for the first time?)'', but the answer ``one-person project Mercury"
is incorrectly translated into ``\Ja{一人称水銀プロジェクト} (the first-person mercury (the chemical element) project)" by Google Translate, which is far from a reasonable answer to the given question.
In this case, because of the lack of the crucial relevant information, the MT system cannot distinguish between the correct translations of ``Mercury'' and ``first-person'' from other words with the same-spelling.
This kind of issue in MT has been studied by previous work in MLQA~\citep{W06-1905,ture-boschee:2016:EMNLP2016}. 
In addition, in extractive RC, the answer spans are sub-phrases in the context paragraph, so \textbf{generating} answer spans by back-translation is not a desirable approach, as it generates homographic variations.
For the baseline on the Japanese SQuAD dataset, we find that only 143 out of 327 answers (44\%) were sub-phrases of the context because of these translation errors or homographic variants.

In contrast with back-translation, our method directly identifies the correct answer in the Japanese context without losing critical relevant information.
It also guarantees that the answers are precisely extracted from the given context.

\subsection{Error Analysis}
We conduct an error analysis of our proposed method on the Japanese and French SQuAD task. 
First, we omit the 41 out of 327 questions (13\%) in the Japanese and French SQuAD datasets where the RC model failed to answer the corresponding questions in the original English SQuAD dataset, so as to focus on the errors that do not stem from the RC.
We randomly sample 100 questions from the remaining 286 paragraph-question pairs, and manually classify errors into three categories: (1) wrong translation of questions, (2) wrong translation of context, and (3) others.
There are 49 errors and 28 errors found respectively in Japanese and French sampled questions.
Some of these errors are caused by multiple factors\footnote{14 cases in Japanese and 6 cases in French}.
Table~\ref{error} shows the error types and the number of the errors.

\begin{table}[t!]
\begin{center}
{\small
\begin{tabular}{|l|c|c|}
\hline &Japanese & French \\\
\bf Type of Errors & \bf \#~(\%)& \bf \#~(\%)\\ \hline
Wrong question translation & 29 (59\%)& 15 (54\%) \\
Wrong context translation & 27 (55\%) & 11 (39\%)\\
Others & 6 (12\%) & 6 (21\%)\\
\hline
\end{tabular}
}
\end{center}
\caption{\label{error} Types of errors and their frequency in the Japanese SQuAD dataset and French SQuAD dataset. }
\end{table}
\paragraph{Wrong Translation of Questions.}

Previous work reports that SQuAD models including BiDAF tend to rely on superficial cues or interrogative of questions~\citep{hu2017mnemonic,zhang2017exploring,jia-liang:2017:EMNLP2017}.
Thus, incorrectly translating a single word in a question sentence can result in a fatal error. 
For instance, the French-to-English NMT model generated the translation ``in the cycle of cycle , that becomes the water when it is heated ?'', when it is given the question ``Dans le cycle de Rankine, que devient l'eau lorsqu'elle est chauffée ? (In the Rankine cycle, what does water turn into when heated?''.
For this case, the RC model failed to find the correct answer `` vapeur (vapour)'', likely because the original interrogative (``what'') is lost.

On the other hand, this implies that even if the questions are not translated appropriately, the keyword-match and heuristics like interrogatives could lead the model to find the correct answers. 
Compared to French-to-English translation, the Japanese-to-English translation tends to produce incorrect translations of questions as shown in Table~\ref{error}.
However, the RC model sometimes nevertheless succeeds in extracting the correct answer spans by leveraging artificial cues.

\paragraph{Wrong Translation of Context.}
The performance of the RC model deteriorates in both languages when the context is incorrectly translated or when parts of the context is missing.
One of the biggest issues of the translation of contexts is the under-translation problem, where some words are mistakenly untranslated.
\citet{tu-EtAl:2016:P16-1} characterize this behaviour for NMT models, which tend to generate shorter translations~\cite{04d503dffe9f4dcfbacf886d1dee56fb} that are valid but miss some phrases.
This is especially common when translating long sentences~\citep{goto-tanaka:2017:NMT}. 
We find that some answer spans were missing or inappropriately translated, which poses significant problems for the RC model. 
For instance, consider a Japanese context ``\Ja{1562年までにユグノーの数はピークに達し、200万人と推定され、フランスの南部及び中部に主に集まり、フランスカトリック協会の構成員の\underline{約8分の1}であった。} (Huguenot numbers peaked near an estimated two million by 1562, concentrated mainly in the southern and central parts of France, \underline{about one-eighth} the number of French Catholics.)'', where the underscored words denote the answer of the question ``What was the proportion of Huguenots to Catholics at their peak?''.
For this paragraph, the MT model generates ``by 1562, the number of huguenots ..., and was estimated to be two million to two million, and was a major gathering of the french catholic society members.''. That is, the expected answer sub-phrase is completely missing. 
This problem is also stated by \citet{lee2018semi}, who observe that while translating the SQuAD training dataset into Korean, the answer spans are often lost in the translated sentence.
We hypothesize that adding explicit constraints to avoid under-translation of important cues should improve performance on the Japanese and French SQuAD datasets.

\paragraph{Others.}
Six errors in the Japanese dataset and six in the French dataset occur due to the lack of robustness of the RC model with respect to paraphrasing. 

\begin{figure}[t]
  \centering
  \includegraphics[width=0.48\textwidth]{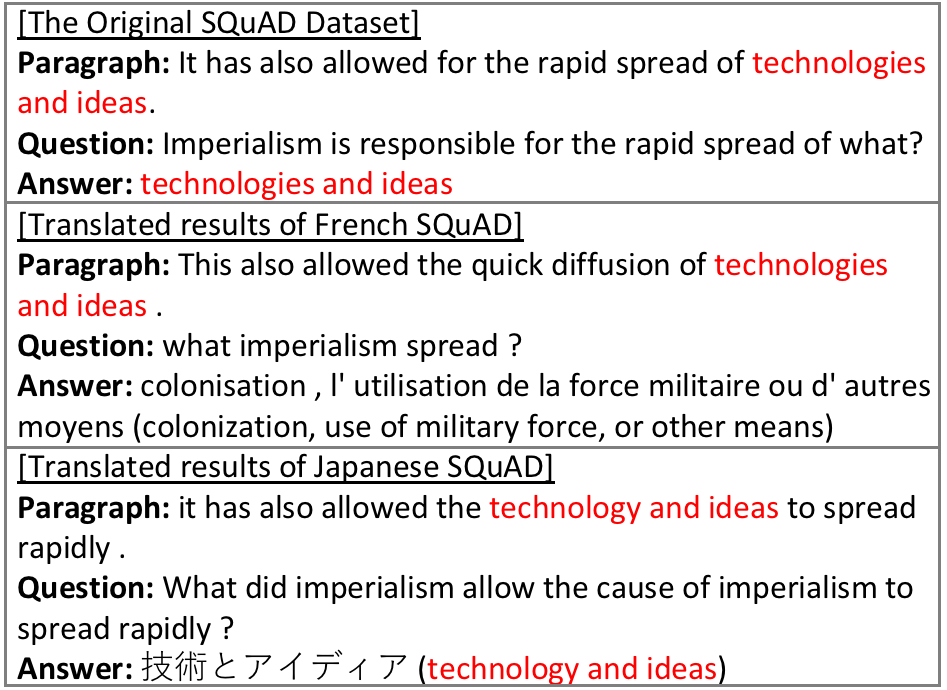}
  \caption{\label{paraphrase} A comparison among the RC results on the English, Japanese, and French SQuAD datasets.}
\end{figure}

Figure~\ref{paraphrase} shows an example of this on the French example. 
Our French-to-English translation model translate the original paragraph and question appropriately, without missing crucial information to answer the given question.
Nevertheless, the subsequent RC model failed because the translation paraphrased a key phrase, ``spread'', to ``diffusion'', in the context.
\citet{weissenborn-wiese-seiffe:2017:CoNLL} report that many questions in SQuAD can be answered with heuristics based on type and keyword-matching, and \citet{jia-liang:2017:EMNLP2017} find that in their adversarial examples experiment focusing on SQuAD, many SQuAD models perform well without being confused with adversarial examples when the question has an exact $n$-gram match with the original paragraph.
These studies suggest that the performance of a SQuAD model relies on the superficial cues, and explains the negative impact on overall RC performance by MT paraphrasing.

\section{Related Work}
\label{sec:related_work}
\paragraph{End-to-end RC.}
End-to-end models have achieved significant performance in extractive RC.
\citet{seo2016bidirectional} proposed the BiDAF network, which represents the context at different levels of granularity and uses a bidirectional attention flow mechanism to obtain a query-aware context representation. 
\citet{wang-EtAl:2017:Long2} used a self-matching attention mechanism to refine the question-aware paragraph representation. 
However, large-scale hand-annotated extractive RC datasets, which are required to train these models, are often exclusive to English. 
Our work proposes a method for leveraging existing English models to produce RC systems for target languages that do not have RC training data.

\paragraph{MLQA}
MLQA is a question answering task in which the questions are formulated in a language different from that of the paragraphs.
Examples of MLQA tasks include QA@CLEF~\footnote{\url{http://www.clef-initiative.eu/track/qaclef}} and Equer~\cite{ayache2006equer}.
A key difference between extractive RC and MLQA is that the questions in extractive RC share the same language as their context, whereas questions in MLQA are formulated in a language different from the context.
Moreover, MLQA contains abstractive answers that must be generated by the model whereas extractive RC answers are spans in the context document.
For these reasons, the approaches in these two tasks are not easily transferable without significant performance degradation.

It is a common approach in MLQA to translate all non-English text or keyword in queries into English beforehand, and then to treat the task as a monolingual  task~\citep{ture-boschee:2016:EMNLP2016,W06-1905,Hartrumpf,esplagomis-sanchezmartinez-forcada:2012:STARSEM-SEMEVAL}.

While both the common MLQA approach and our work involve combining MT with question answering, the goal of the tasks are distinct.
The former emphasizes joint reasoning across language while the latter emphasizes building practical RC systems for target languages without any training data.

\paragraph{Datasets for non-English RC}
\citet{lee2018semi} proposed a method to create a large-scale training dataset for Korean by translating an existing large-scale English RC dataset into Korean, adding a few thousand manually annotated Korean paragraph-question pairs. 
\citet{he2017dureader} created DuReader, a new large-scale open-domain Chinese RC dataset. 
These works focus on training an RC model in a language of interest by creating large-scale datasets for the languages manually or semi-automatically.
These approaches require additional annotations of RC datasets, which might be costly, and do not leverage the wealth of exiting resources in English RC.
In contrast, we propose a method to build an RC system without using any additional RC training data in the language of interest.
Furthermore, our technique leverages existing resources in English RC.

\section{Conclusion}
We proposed an RC system for a target language with no RC training dataset by combining existing English RC models with an attentive NMT model, using soft-alignments from the latter to recover answers in the target language.
Our results showed that our approach significantly outperforms a back-translation method with a state-of-the-art MT system on the Japanese and French SQuAD task. 
In future work, we will investigate how to improve system robustness toward paraphrasing, and how to alleviate the problem of missing key phrases that stem from NMT.

\section*{Acknowledgments}
% \todo{remove for NAACL}
We thank Victor Zhong for his help in revising the paper.
This project is partially financially supported by Microsoft Japan.
This work is also partially supported by JST CREST Grant Number
JPMJCR1513, Japan.

\bibliography{naaclhlt2019}
\bibliographystyle{acl_natbib}

\bigskip

\clearpage
\newpage	

\appendix

\section*{Appendix}

\subsection*{A. Details of Wikipedia-based Bilingual Corpus Creation}
\label{sec:bilingual_corpus}
To enable our system to translate paragraphs and questions on the wide range of topics, we train our NMT model with a large Wikipedia bilingual corpus, which is created in the following steps. 

First, we collected Japanese, French and English Wikipedia dump files and extracted the plain texts from each article.
For a Japanese-to-English Wikipedia-based corpus creation, we used the Japanese and English Wikipedia page articles dataset dumped in December, 2017\footnote{\url{https://dumps.wikimedia.org/jawiki/latest/jawiki-latest-pages-articles.xml.bz2-rss.xml}}\footnote{\url{https://dumps.wikimedia.org/enwiki/latest/enwiki-latest-pages-articles.xml.bz2-rss.xml}}, containing 1,085,986 Japanese Wikipedia articles and 5,523,723 English Wikipedia articles. 
For the creation of a French-to-English Wikipedia-based corpus, we used the French Wikipedia pages article dataset dumped in April, 2018\footnote{\url{https://dumps.wikimedia.org/frwiki/20180420/frwiki-20180420-pages-meta-current.xml.bz2}}, consisting of 1,976,603 French Wikipedia articles, and the same English Wikipedia articles used in the Japanese-to-English corpus creation.

We extracted the plain text of each article using an Wikipedia article extraction tool\footnote{\url{https://github.com/attardi/wikiextractor}}. 
After  collecting the plain text data, 
we utilized the latest langlinks data of Japanese and French Wikipedia data\footnote{\url{https://dumps.wikimedia.org/jawiki/latest/jawiki-latest-langlinks.sql.gz}}\footnote{\url{https://dumps.wikimedia.org/frwiki/20180420/frwiki-20180420-langlinks.sql.gz}}, which stores a list of all interlanguage links from the provided pages to other languages. 
There are 479,551 and 1,525,465 Wikipedia articles in Japanese and French, respectively (around 44\% and 77\% of the respective entire Wikipedia articles).
The articles have their interlanguage links to English Wikipedia articles as well.
For French, 1,525,465 French Wikipedia articles (around 77\% of the entire French Wikipedia articles) have their interlanguage links to English Wikipedia articles.
Then we aligned the sentences in Japanese and French articles to the sentences in English articles by using a sentence-level alignment tool. 
We employed the hunalign sentence aligner\footnote{\url{https://github.com/danielvarga/hunalign}}, which aligns bilingual text on the sentence level, based on the sentences' length and dictionary-based translations~\citep{varga2007parallel}. 
Although the aligner does not handle changes of sentence order, we assume that the sentence order are less likely to be changed among different language articles on Wikipedia. 
Hunalign takes a bilingual phrase dictionary, and translates sentences in a source language to a target language based on the phrase dictionary, so that it could calculate a similarity score between a pair of sentences.
For the Japanese-to-English dictionary-based translation process, we used MUSE's ground-truth bilingual dictionaries of Japanese and English\footnote{\url{https://github.com/facebookresearch/MUSE}}, and the EDICT Dictionary File\footnote{\url{http://www.edrdg.org/jmdict/edict.html}}, a Japanese-English Dictionary file containing about 175,000 entries.
Regarding French-to-English bilingual phrase dictionary, we used MUSE's ground-truth bilingual dictionaries of French and English\footnote{\url{https://github.com/facebookresearch/MUSE}}.
We omitted the sentence pairs whose alignment score is lower than -0.3 and 0.0 for Japanese-to-English and French-to-English, respectively.
We also filtered out the translation pairs whose sentence lengths are longer than 50 or shorter than 5.
As a result, we collected 4,567,800 sentence pairs for Japanese-to-English, and 6,398,489 sentence pairs for French-to-English.
We extracted the best aligned 1,002,000 pairs to build our \{Japanese, French\}-to-English NMT training and development datasets.

\subsection*{B. Details of Manually Translated SQuAD Dataset Questions Creation}
To obtain Japanese and French translation of randomly sampled 200 question sentences, we used Amazon Mechanical Turk\footnote{\url{https://www.mturk.com/git}{}} and asked bilingual workers to translate the questions in English into the target languages (Japanese and French) accurately without using any translation software or web service such as Google Translate. 
We assigned 20 question sentences to each worker.

\subsection*{C. Details of Experimental Settings of MT}
% \subsection{Experimental Settings}
We use the Wikipedia bilingual corpora introduced in Section~\ref{subsec:wiki_enja} for training \{Japanese, French\}-to-English NMT models.
The word embeddings and the weight matrices of the NMT model are initialized uniformly with random values in $[-0.1,+0.1]$. 
We train using batched stochastic gradient descent with a batch size of 128, a momentum of 0.75, and an initial learning rate of 1.0.
We used 512-dimensional hidden states and embeddings.
We use gradient clipping with a threshold of 1.0~\citep{Pascanu:2013:DTR:3042817.3043083}.
We calculate the BLEU score~\citep{papineni-EtAl:2002:ACL} of greedy translations on each development data set at every half epoch while training the models.
We use L2-norm regularization with a coefficient of $10^{-6}$ and applied dropout \citep{hinton2012improving} with a dropout rate of 0.2. 
We build the vocabulary with words appearing more than five times in the corpus.
When oversampling the questions introduced in Section~\ref{subsec:question_translation}, we set the duplication factor of the questions to $l=10$.
At the test time, we used a beam search method proposed in~\citet{wat_oda}, with a beam size of 5.

\subsection*{D. Pre-Processing for Japanese and French Sentences}
For Japanese sentences in the NMT training dataset and SQuAD test dataset, we first normalized sentences with Normalization Form Compatibility Composition (NFKC) and tokenized the sentences by {\it MeCab}\footnote{\url{http://taku910.github.io/mecab/}{}}. 
For French, we normalized punctuation by {\it moses-SMT's punctuation normalizing script}\footnote{\url{https://github.com/moses-smt/mosesdecoder/blob/master/scripts/tokenizer/normalize-punctuation.perl}{}}, and tokenized the sentences by {\it moses-SMT's tokenizer}\footnote{\url{https://github.com/moses-smt/mosesdecoder/blob/master/scripts/tokenizer/tokenizer.perl}{}}.

\subsection*{E. Details of Experimental Settings of RC}
We train BiDAF~\citep{seo2016bidirectional} and BiDAF + Self Attention + ELMo model~\citep{clark2017simple} with the original SQuAD v1.1 English training dataset.
For the training of the BiDAF model, we follow the training settings in \citet{seo2016bidirectional}, except that we set the batch size to 40 and trained the model for 20 epochs.
In case of training the BiDAF + Self Attention + ELMo model, we follow the settings shown in \citet{clark2017simple}, adding ELMo embeddings~\cite{peters2018deep}.
The only distinction is that we use 100 dimensional pre-trained GloVe~\cite{pennington2014glove}, instead of the 300 dimensional one. 
To evaluate the performance, we calculate EM, which measures the exact match with the ground truth answers, and F1, the weighted average of the precision and recall rate at character level~\citep{D16-1264}.
The initial learning rate is set to 0.5, and the learning rate was halved when no improvement was seen in the EM score during two epochs.

\subsection*{F. Comparison to translating the entire pivot language dataset}
Contrary to our approach, which translates the target language into the pivot language, one can alternatively translate the existing large-scale dataset from the pivot language $P$ to the language $L$, and subsequently train an RC model on the large translated dataset to build a RC model in language $L$.
Transferring English resources to the target language using MT has been proposed for a variety of NLP tasks~\cite{tiedemann-agic-nivre:2014:W14-16,lee2018semi,balahur2012comparative}, and is one of the common approach in multi-lingual NLP.
Yet, considering the nature of extractive RC, transferring the existing large-scale datasets into the target language has several obstacles:

\begin{itemize}
\item in extractive RC, the answer must be a continuous sub-span of the target language context $C_L$, and thus small homographic variations introduced by translation prevent the model from finding the correct answer.
\item a long context in the pivot language $C_P$ is likely to result in shorter translated context $C_L$ due to the fact that NMT models tends to generate shorter translations~\cite{04d503dffe9f4dcfbacf886d1dee56fb}, missing some important information in the original languages~\citep{goto-tanaka:2017:NMT}. 
This may result in the answer being missing in the target language $L$.
\end{itemize}

We sampled 100 paragraph-question pairs with ground-truth answers from the SQuAD training set, translated the sampled pairs into Japanese using Google Translate\footnote{The translations were obtained at \url{https://translate.google.com} in October, 2018.}, and checked how many of the translated answers were actually preserved in the translated paragraph. 
We observed that for 51 out of the 100 questions (51\%), the translated answer did not match any spans in the translated context. 
The main reasons for this are that the answer spans tend to be lost at the translation process (14 out of 51).
Even if the answers are preserved, small variants between translated answers and the spans in the translated paragraph occurred (36 out of 51). 
% The details of the errors can be found in the supplementary material Section A.
% \todo{fix the detailed explanation about the appendix or remove it}

In addition to our finding,~\citet{lee2018semi} also report that training a RC model only on automatically translated SQuAD training dataset resulted in poor performance because of translation errors. 
Therefore, we conclude that translating the existing training data from the pivot language $P$ into the target language $L$ to train a RC model in language $L$ would fail to achieve good performance.

\end{document}